\documentclass[lettersize,journal]{IEEEtran}
\usepackage{amsfonts}
\usepackage{algorithmic}
\usepackage{algorithm}
\usepackage{array}
\usepackage[caption=false,font=normalsize,labelfont=sf,textfont=sf]{subfig}
\usepackage{textcomp}
\usepackage{stfloats}
\usepackage{url}
\usepackage{verbatim}
\usepackage{cite}

\usepackage{multirow}
\usepackage{listings}
\usepackage{orcidlink}
\usepackage{cite}

\usepackage[cmex10]{amsmath}
\usepackage{url}

\usepackage{color}
\usepackage{etoolbox}
\usepackage{setspace}

\begin{document}
\title{SWAN: Self-supervised Wavelet Neural Network for Hyperspectral Image Unmixing}

\author{Yassh~Ramchandani,~Vijayashekhar~S~S,~and~Jignesh~S.~Bhatt%
\thanks{Yassh Ramchandani is Lead AI Engineer at Info Edge India Ltd, Noida-201317, India (e-mail: yasshramchandani@gmail.com).

Vijayashekhar S S is with the Department of Artificial Intelligence and Machine Learning, Acharya Institute of Technology, Bengaluru-560107, India (e-mail: vijayashekar\_2221@acharya.ac.in).

Jignesh S. Bhatt is with the Indian Institute of Information Technology Vadodara (IIITV), c/o Block 9, Government Engineering College, Sector 28, Gandhinagar-382028, India (e-mail: jignesh.bhatt@iiitvadodara.ac.in).}}%

\markboth{Ramchandani \MakeLowercase{\textit{et al.}}: SWAN: Self-supervised Wavelet Neural Network for Hyperspectral Image Unmixing}{}
\maketitle

\begin{abstract}
Hyperspectral image unmixing is an ill-posed inverse problem. In this article, we present SWAN: a three-stage, self-supervised wavelet neural network for joint estimation of endmembers and abundances from hyperspectral imagery. The contiguous and overlapping hyperspectral band images are first expanded to Biorthogonal wavelet basis space that provides sparse, distributed, and multi-scale representations. The idea is to exploit latent symmetries from thus obtained invariant and covariant features using self-supervised learning paradigm. The first stage, SWANencoder comprises five fully-connected layers that map input wavelet coefficients to a compact lower-dimensional latent space. The second stage, SWANdecoder forms two parallel layers and uses the derived latent representation to reconstruct the input wavelet coefficients. Interestingly, the third stage SWANforward learns the underlying physics of the hyperspectral image using four fully-connected layers. A three-stage combined loss function is formulated in the image acquisition domain that eliminates the need for ground truth and enables self-supervised training. Adam is employed for optimizing the proposed loss function while Sigmoid with a dropout of 0.3 is incorporated to avoid possible overfitting. Kernel regularizers bound the magnitudes and preserve spatial variations in the estimated endmember coefficients. The output of SWANencoder represents estimated abundance maps during inference, while weights of SWANdecoder are retrieved to extract endmembers. Experiments are conducted on two benchmark synthetic data sets with different signal-to-noise ratios (SNRs) as well as on three real benchmark hyperspectral data sets while comparing the results with several state-of-the-art neural network-based unmixing methods. The qualitative, quantitative, and ablation results show performance enhancement by learning a resilient unmixing function as well as promoting self-supervision and compact network parameters for practical applications.
\end{abstract}


\IEEEpeerreviewmaketitle

\newcolumntype{M}[1]{>{\centering\arraybackslash}m{#1}}

\section{Introduction and Literature Review}
Hyperspectral imaging remotely senses a geographical location in a range of wavelengths with a high spectral resolution, typically $10 nm$. This enables the identification of materials based on their reflective spectral features called signatures. Such hyperspectral images often come at the cost of a poor spatial (ground) resolution, typically $1m^{2} \text{ to } 30 m^{2}$, due to which there is usually more than one material found in every pixel of the acquired imagery. Spectral unmixing reveals the scene characteristics in terms of extracting the constitutional signature values of spectrally distinct materials called endmembers and estimating corresponding fractional contributions by each endmember at each pixel location known as abundance (material) maps. 

For more than three decades, researchers have presented numerous techniques for spectral unmixing using linear and non-linear mixing data models. The popular and well-accepted linear mixing models (LMMs) assume that spectral mixing takes place on a macroscopic scale and that the interaction between incident light and different materials happens individually. The non-linear mixing models assume more intricate physical interactions between the light scattered by many different materials in a scene, where the interactions are at a microscopic level \cite{heylen2014review, dobigeon2013nonlinear}. The problem of spectral unmixing is interpreted as a geometrical, statistical, sparse regression, or deep learning-based problem, leading to four broad categories of unmixing methods \cite{bioucas2012hyperspectral, wang2019hyperspectral, signoroni2019deep, jignesh2020deep, rasti2020feature, palsson2022blind}.

The geometrical-based methods can be categorized into pure pixel or minimum volume methods. A few prime examples are vertex component analysis (VCA) \cite{nascimento2005vertex}, and minimum volume simplex analysis (MVSA) \cite{li2015minimum}. The statistical methods formulate the unmixing problem as an inference problem \cite{nascimento2011hyperspectral, dobigeon2009joint, eches2010bayesian, bhatt2014data}. Sparse regression methods assume that the observed spectra can be expressed as linear combinations of known spectral signatures from available digital spectral libraries \cite{iordache2011sparse,iordache2012total, iordache2013collaborative, iordache2013music}. Methods based on compressed sensing \cite{zhang2014structured, albayrak2014compressed, xusu2016compressive} also belong to this category.

With the advent of artificial intelligence technology, researchers in the last decade have applied deep learning techniques and obtained encouraging results for hyperspectral image unmixing (HSIU). The challenges and possible opportunities for addressing HSIU from the deep learning perspective are summarized in \cite{wang2019hyperspectral, signoroni2019deep, jignesh2020deep, rasti2020feature, palsson2022blind}. It is found that autoencoder, convolutional neural network, generative model, and transformer-based architectures are prevalent for HSIU. One of the pioneer examples of using an autoencoder neural network with hyperspectral data is demonstrated in \cite{licciardi2018spectral}. Initially, the autoencoder architecture has been widely used and studied for HSIU applications \cite{palsson2018hyperspectral, wang2019nonlinear, palsson2022blind}. Authors in \cite{endnet} have demonstrated a three-staged autoencoder network with a novel loss function to improve the sparsity of the estimates, while authors in \cite{qu2018udas} propose an untied denoising autoencoder with sparsity (uDAS) for spectral unmixing. Followed by this, \cite{su2019daen} addresses the issue of outliers by using a stacked autoencoder to learn the spectral signatures followed by a variational autoencoder (VAE) to jointly estimate the endmembers and abundances. In contrast, \cite{egu_net} extends the autoencoder architecture to a deep Siamese network, which uses the pure or nearly-pure endmembers to mutually learn and improve shared model weights while adding spectrally meaningful constraints. Authors in \cite{min2021jmnet} consider the features in the observed and reconstructed data in order to regularize the conventional autoencoder training using Wassertein distance and feature matching. Meanwhile, a deep generative endmember model is trained using pure pixel information with a VAE in \cite{borsoi2019deep}. One of the limitations of autoencoder based approaches is that they process a single spectrum at a time and hence inherently exclude spatial context. Nevertheless, the autoencoder based unmixing solution is regularized by incorporating spectral and spatial priors in \cite{patel2021spectral}. The method referred to as unmixing using deep image prior (UnDIP) \cite{rasti2022undip} utilizes endmembers extracted by a simplex volume maximization (SiVM) technique. Recently, a minimum simplex convolutional network (MiSiCNet) \cite{rasti2022misicnet} is proposed to incorporate both the spatial correlation between adjacent pixels and the geometrical properties of the linear simplex. Further, authors in \cite{zouaoui2023edaa} employ entropic gradient descent startegy to traditional archetypal analysis (EDAA) to obtain better solutions to blind unmixing problem, specifically to allow efficient GPU implementation for the same.

It is known that deep convolution neural networks (CNNs) have the ability to build large-scale invariants which are stable to spatial deformations \cite{lecun2010convolutional}. Therefore, to incorporate both the spectral and spatial features in the network, authors in \cite{zhang2018hyperspectral, palsson2020convolutional,deep_spectral_convnet} use spectral features with a convolution network. It is then improved in \cite{ozkan2018improved} by combining the endmember uncertainty with multinomial mixture kernel. It employs Wasserstein generative adversarial network (WGAN) to improve stability during the network optimization and capture the uncertainty. Meanwhile, authors in \cite{zhao2021hyperspectral} present a 3-D CNN model to exploit the spectral-spatial structures as well as the non-linear features in the data. Authors in \cite{ghosh2022hyperspectral} posit that performing HSIU with a transformer network captures non-local feature dependencies by interactions among image patches, which are not employed in a typical CNN model.

To ensure a deep network's performance is unbiased and generalized, there is a need for high quality and high quantity data sets for training and test. Authors in \cite{wu2018semi} use semi-supervised deep-learning for HSI classification using convolutional recurrent neural networks, while authors in \cite{cheng2024safdnet}  propose a deep-network architecture based on stochastic adaptive Fourier decomposition (SAFD) theory to train a HSI classifier with small number of annotated images. We recently presented a self-supervised learning-based approach for HSIU to overcome the issue of limited availability of ground truth imagery while better handling the noise and perturbations in the acquired images \cite{ssl_hsu}. A comprehensive review on self-supervised learning for computer vision in the context of remote sensing can be found in \cite{wang2022sslreview}. 

Interestingly, a few recent works found that representing the acquired image in a suitable transformed domain enables a learning machine to perform significantly better with lesser computations for tasks involving hyperspectral image processing \cite{chenot2018blind,vijay2019novelapproach, xu2020curvelet, chakraborty2021spectralnet, guo2022review, cheng2024safdnet}. Authors in \cite{wavelet_autoencoder} primarily use a vanilla autoencoder with Kullback–Leibler divergence-based minimization for unmixing the hyperspectral data. In \cite{blind_hsu}, we unveil the compressibility of spectrally dense and overlapped hyperspectral images by developing a compact linear mixing model in the wavelet domain.

In this article, we propose SWAN: a three-stage, self-supervised, wavelet-featured neural network to better handle the ill-posedness of blind HSIU. The network leverages sparse wavelet coefficients that capture local variations at different scales to effectively learn the unmixing function with a very few hidden parameters. Besides inverse-forward stages, a separate stage is included to learn the physics of hyperspectral images and increase the network's resilience to implicit noise. A three-stage combined loss function is formulated in the measurement (acquisition) domain and hence eliminates the need for ground truth unmixed components during training. Custom regularizers are employed to further make the problem better-posed and reinforce the enhanced representation of hyperspectral imagery.

The key contributions of this article include:
\begin{enumerate}
    \item A compact learning machine in wavelet space for joint estimation of endmembers and abundances.
    \item Self-supervised learning of unmixing function using Biorthogonal wavelet basis.
    \item Stable end-to-end network training in the absence of the ground truth unmixed image components and with high implicit Gaussian noise in the input images. 
\end{enumerate}

\newcommand{\rom}[1]{\uppercase\expandafter{\romannumeral #1\relax}}

The rest of this article is structured as follows: Section \rom{2} describes the problem set-up. In Section \rom{3}, the proposed SWAN network architecture is discussed. Experimental methodology and results are discussed in Section \rom{4}. Finally, conclusions drawn from this research work, and application scenario are presented in Section \rom{5}.

\section{Problem Set-up and Insights}
\begin{figure}[h]
\centering
\includegraphics[width=\linewidth]{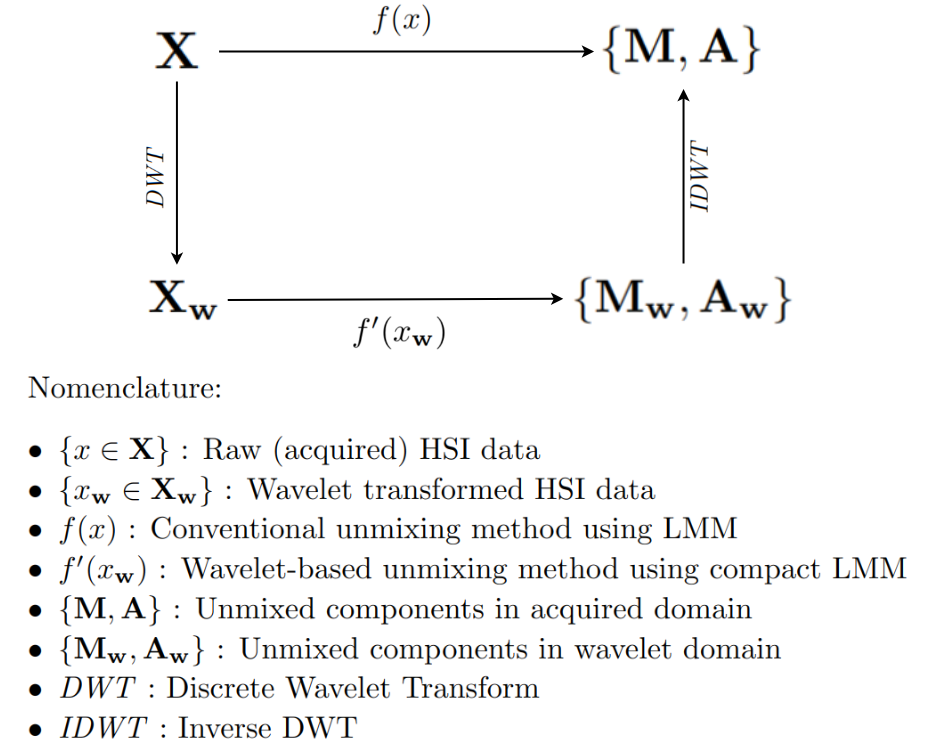}
\caption{Function map of problem set-up}
\label{function_map}
\end{figure}

Referring to Fig. \ref{function_map}, a generic deep learning approach learns an unmixing function $f(x)$ directly in the data domain, given the hyperspectral image data $\mathbf{X}$ perturbed with the Gaussian noise $\mathbf{N}$; and yields estimated endmembers $\mathbf{\widehat{M}}$ and abundnace maps $\mathbf{\widehat{A}}$ under the LMM with \textit{e} number of endmembers, 
\begin{equation} \label{linear_mm}
    \mathbf{X} = \mathbf{M\;A} + \mathbf{N}\:.
\end{equation}

In this work (Fig. \ref{function_map}), we propose the acquired hyperspectral image $\mathbf{X}_{(L\times P)}$ is expanded (represented) into wavelet domain as $\mathbf{X_w}_{(K\times P)}$ by applying discrete wavelet transform (DWT) \cite{mallat1999wavelet} to each hyperspectral image vector, i.e.
\begin{equation}\label{dwt_tranform}
\mathbf{X_w}_{(K\times P)} = \{\mathbf{X_a, X_d}\} = \{DWT(\mathbf{x}_{i(L\times 1)})\}_{i=1}^P \:. 
\end{equation}

Here \textit{P} and \textit{L} represent the number of pixels and bands in a hyperspectral image data cube, respectively, while \textit{K} is the number of bands in a hyperspectral image cube in the wavelet domain. The $\mathbf{X_a \text{ and } X_d}$ capture the gross (approximate) and finer (detail) symmetries from the hyperspectral band images, respectively. See that (Fig. \ref{function_map}), our work learns $f'(x_{\mathbf{w}})$ from $\mathbf{X_{w}}$ and better estimates the unknowns $\{\mathbf{M\text{, }A}\}$. \textit{We posit that such a representation provide a more aligned set of features to better facilitate a neural network to handle the ill-posedness of the unmixing}. Our recent work \cite{blind_hsu} demonstrates the compact representational ability of biorthogonal family of wavelets, and hence we rewrite the compact LMM in the wavelet domain as in \cite{blind_hsu}:
\begin{equation}\label{lmm_wavelet_domain}
    \{\mathbf{X_a}, \mathbf{X_d}\}_{K\times P} = \{\mathbf{M_a}, \mathbf{M_d}\}_{K\times e}\; \mathbf{A}_{e\times P} + \mathbf{N}_{K\times P} \:,
\end{equation}
where \textbf{a} and \textbf{d} denote set of approximation and detail coefficients of endmember matrices \{$\mathbf{M_a}, \mathbf{M_d}$\} in wavelet domain. 

\textbf{Problem definition:} Given \textit{P} number of \textit{K}-dimensional hyperspectral image vectors in wavelet space, our objective is to construct a self-supervised wavelet neural network (SWAN) for joint estimation of endmembers and abundance maps.

\textbf{Motivation:} To learn underlying unmixing function $f(x)$ from hyperspectral imagery, typically a neural network progressively contracts the input space and linearizes transformations in the directions along which the unknown function remains nearly constant \cite{tishby2015bottleneck, mallat16, glass_box_cnn}. These directions correspond to the groups of local symmetries that better condition the unknown function to be invariant or covariant to specific features in the images. One of the ways for the unmixing of spectrally dense and overlapped hyperspectral imagery is to learn these latent features which are efficiently captured in the multi-scale, sparse wavelet coefficients \cite{blind_hsu, glass_box_cnn, wavelet_autoencoder, chakraborty2021spectralnet, manaswini2024analyticalcnn}. When presented with hierarchical non-linearities, the multi-scale and sparse wavelet transformed representation facilitates a neural network to effectively learn the unmixing function $f'(x_{\mathbf{w}})$ (Fig. \ref{function_map}) while preserving the invariances and covariances to the latent spectral features. It motivates us to construct SWAN architecture based on self-supervised learning to perform blind HSIU in the wavelet domain.

\textbf{Advantages:} Following are specific benefits of the proposed work: (i) eliminating the need for ground truth unmixed components (self-supervised learning), (ii) learning the unmixing function $f'(x_{\mathbf{w}})$ with a very few hidden parameters ($\approx$50,000 weights using wavelet representation), (iii) having lesser computational complexity w.r.t. plain neural network for HSIU, (iv) achieving stable training, and (v) better addressing the representational aspects of rich spatio-spectral information available in the wavelet space of hyperspectral images. Together, these shall better handle the ill-posedness of learning unmixing function and help making the problem better-posed.

\section{Proposed SWAN}

\begin{figure*}[t]
\centering
\includegraphics[width=\textwidth, height=9cm]{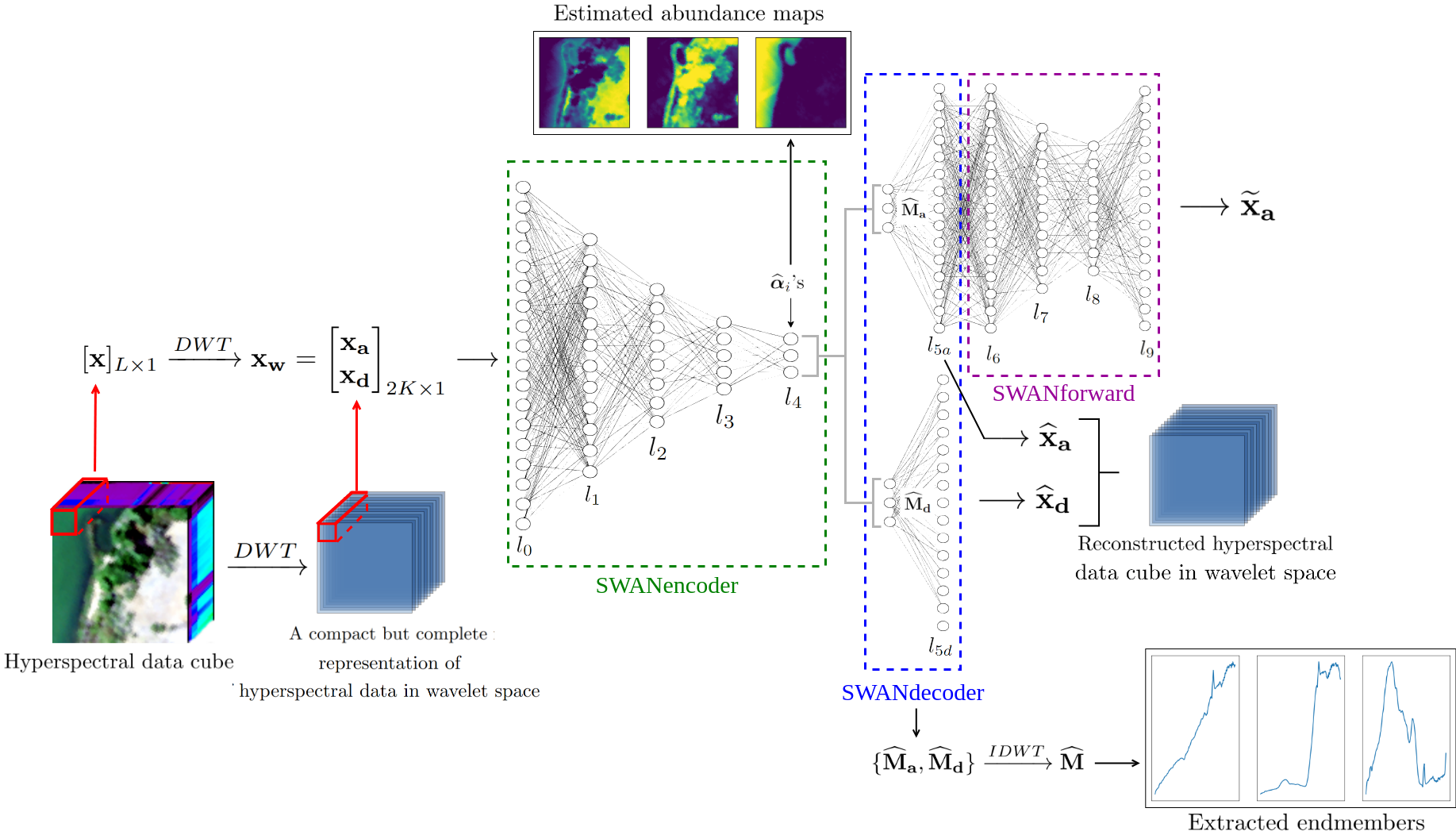}
\caption{Schematic diagram of the proposed SWAN for hyperspectral image unmixing.}
\label{proposed_network_architecture}
\end{figure*}

\begin{figure*}[t]
\centering
\includegraphics[width=\textwidth, height=7.5cm]{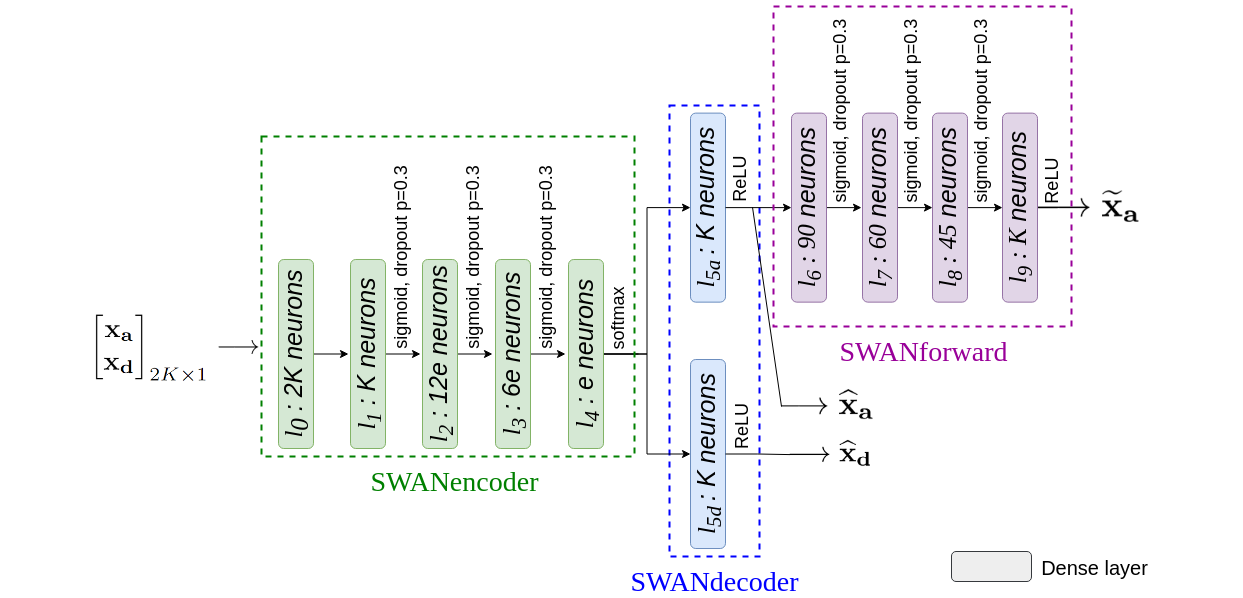}
\caption{Detailed neural architecture design of the proposed SWAN in Fig. \ref{proposed_network_architecture}.}
\label{detailed_network_architecture}
\end{figure*}

The joint unmixing problem defined in section \rom{2} requires the proposed SWAN architecture to address three main objectives, viz. (a) joint estimation of endmembers and abundance maps of hyperspectral images, (b) use of self-supervised learning paradigm, and (c) neural information processing based on wavelet representations. To achieve these objectives, we design a three-stage network structure that learns to efficiently compress and then reconstruct the hyperspectral images by decomposing them to perform blind hyperspectral image unmixing. Interestingly, the network's training objective (loss) function is formulated in the domain of image acquisition to facilitate self-supervised learning and incorporates $l_1\text{ and }l_2$ regularizers to better handle the ill-posedness of HSIU problem. In this section, we present the detailed structure of the three SWAN sub-networks - SWANencoder, SWANdecoder, and SWANforward, the network's training objectives, and the inference methods used to extract the estimated endmembers and abundance maps.

Fig. \ref{proposed_network_architecture} shows schematic details of the proposed SWAN for hyperspectral image unmixing. Given a hyperspectral data vector $\mathbf{x}$ of dimension $L \times 1$ at a single pixel location in hyperspectral image cube, we obtain a set of approximation and detail coefficients $\{\mathbf{x_a}, \mathbf{x_d}\} = DWT(\mathbf{x})$ using Eq. (\ref{dwt_tranform}). These coefficients of dimensions $2K \times 1$ serve as input to SWAN. As shown in Fig. \ref{proposed_network_architecture} the three-staged, self-supervised architecture comprises an encoder, a decoder, and a forward model. For each training batch as the input, SWAN reconstructs the estimated image vector and the denoised image vector at each pixel during the forward pass while backpropagates the error to learn an optimal unmixing function (Eq. (\ref{lmm_wavelet_domain})). During inference, we use the encoder's output and the connected weights of the decoder to construct the estimated abundance matrix $\mathbf{\widehat{A}}$ and the estimated endmember matrix $\mathbf{\widehat{M}}$, respectively. Fig. \ref{detailed_network_architecture} shows the detailed neural architecture design for the proposed SWAN shown in Fig. \ref{proposed_network_architecture}. Each cell in Fig. \ref{detailed_network_architecture} represents a fully connected dense layer with respective details about the number of neurons, activation function, and dropout mask mnetioned at each layer's output. We shall now discuss each stage in detail, including the training and inference phases.

Referring to Fig. \ref{proposed_network_architecture}, the SWANencoder is an undercomplete encoder that maps the input hyperspectral wavelet coefficients to a lower-dimensional latent space. Note that the wavelet representation of HSI facilitates exploiting symmetries, invariant, and covariant features \cite{manaswini2024analyticalcnn}. As shown in the detailed architecture Fig, \ref{detailed_network_architecture}, the SWANencoder consists of five fully connected layers $l_0 \text{ to } l_4$. Sigmoid activation with a dropout of 0.3 is employed at $l_1, l_2, l_3$ layer's output to introduce non-linearity and ensure that all the neurons remain significantly active in each layer, respectively. Softmax activation at $l_4$ layer's output incorporates the sum-to-one constraint on estimated abundances.

Then, the complementary SWANdecoder uses the learned latent representation and reconstructs the approximation and detail wavelet coefficients. In order to reconstruct them separately, the SWANdecoder consists of two parallel layers $l_{5a}$ and $l_{5d}$ with ReLU activation as shown in Figs. \ref{proposed_network_architecture} and \ref{detailed_network_architecture}.

Finally, the SWANforward (Figs. \ref{proposed_network_architecture} and \ref{detailed_network_architecture}) comprises a series of fully connected layers $l_6 \text{ to } l_9$ that learn the underlying physics of data acquisition and in turn make the network resilient to high implicit noise. Sigmoid activation with a dropout of 0.3 is employed at $l_6, l_7, l_8$ layer's output, while ReLU is used for layer $l_9$. Note that the SWANforward is connected with layer $l_{5a}$ only. Here. we rely on the practical observation that most of the information is captured by the approximation wavelet coefficients of the data under normality assumptions \cite{mallat1999wavelet}. 

As shown in Fig. \ref{proposed_network_architecture}, the SWANdecoder's output consists of the estimated set of denoised approximation and detail coefficients $\{\mathbf{\widehat{x}}_a, \mathbf{\widehat{x}}_d\}$, while the SWANforward's output is the estimated set of approximation coefficients $\mathbf{\widetilde{x}_a}$. These outputs will be compared with the input using a combination of mean-squared error (MSE) \cite{rmse_cite} and spectral angle distance (SAD) \cite{sad_cite}. The combined loss takes care of errors in both the magnitudes and angles of the estimated hyperspectral image data vectors. The proposed three-stage loss function for SWAN is as follows:
\begin{align}
\label{loss_function}
    \mathcal{L} = \mathcal{L}_{5a}(\mathbf{x_a}, (\mathbf{\widehat{x}_a+n})) + \mathcal{L}_{5d}(\mathbf{x_d}, \mathbf{\widehat{x}_d}) + \mathcal{L}_{9}(\mathbf{x_a}, \mathbf{\widetilde{x}_a})\; + \\ \nonumber
    \lambda_1 ||\mathbf{W}^{(l_{5a})}||_2 + \lambda_2 ||\mathbf{W}^{(l_{5d})}||_1
\end{align}
where $\mathbf{n}\in \mathbb{R}^{K\times 1}$ is the Gaussian noise vector, while $\lambda_1$ and $\lambda_2$ are the regularization parameters derived using cross-validation technique \cite{bhatt2016regularization}. Here, we use Gaussian noise as a prior to extract the denoised approximation coefficient $\mathbf{\widehat{x}_a}$. Note that the model training is self-supervised in nature since the input imagery itself serves as the supervision during the training.

In Eq. (\ref{loss_function}), $\mathcal{L}_{5a}, \mathcal{L}_{5d}$ and $\mathcal{L}_9$ are defined for a batch size of $\mathcal{B}$  as follows:

\begin{multline}
\label{l5a}
    \mathcal{L}_{5a}(\mathbf{x_a}, (\mathbf{\widehat{x}_a+n})) = \frac{1}{\mathcal{B}} \sum_{i=1}^{\mathcal{B}}||\mathbf{x_a}_i-(\mathbf{\widehat{x}_a+n})_i||_2^2\; + \\ \frac{1}{\mathcal{B}}\sum_{i=1}^{\mathcal{B}}\arccos\left( \frac{\langle \mathbf{x_a}_i,(\mathbf{\widehat{x}_a+n})_i\rangle}{||\mathbf{x_a}_i||_2\;  ||(\mathbf{\widehat{x}_a+n})_i||_2}\right)\;,
\end{multline}

\begin{multline}
\label{l5d}
    \mathcal{L}_{5d}(\mathbf{x_d}, \mathbf{\widehat{x}_d}) = \frac{1}{\mathcal{B}}\sum_{i=1}^{\mathcal{B}}\arccos\left( \frac{\langle \mathbf{ x}_{\mathbf{d}i},\mathbf{\widehat{x}}_{\mathbf{d}i}\rangle}{||\mathbf{x}_{\mathbf{d}i}||_2\;  ||\mathbf{\widehat{x}}_{\mathbf{d}i}||_2}\right)\;,
\end{multline}

\begin{multline}
\label{l9}
    \mathcal{L}_9(\mathbf{x_{a}}, \mathbf{\widetilde{x}_{a}}) = \frac{1}{\mathcal{B}} \sum_{i=1}^{\mathcal{B}}||\mathbf{x}_{\mathbf{a}i}-\mathbf{\widetilde{x}}_{\mathbf{a}i}||_2^2\; + \\ \frac{1}{\mathcal{B}}\sum_{i=1}^{\mathcal{B}}\arccos\left( \frac{\langle \mathbf{x}_{\mathbf{a}i},\mathbf{\widetilde{x}}_{\mathbf{a}i}\rangle}{||\mathbf{x}_{\mathbf{a}i}||_2\;  ||\mathbf{\widetilde{x}}_{\mathbf{a}i}||_2}\right)\;.
\end{multline}

Referring to Eqs. (\ref{loss_function} - \ref{l9}), the $\mathcal{L}_{5a}$ and $\mathcal{L}_{5d}$ loss terms enable effective learning of an unmixing function by SWANdecoder, while $\mathcal{L}_{9}$ loss term ensures robustness to high noise levels in the input images. The $l_2$-regularizer on $\mathbf{W}^{(l_{5a})}$ better handles the ill-posedness of the problem by constraining the bounds on the magnitudes of approximation coefficients, while the $l_1$-regularizer on $\mathbf{W}^{(l_{5d})}$ preserves the spatial variations in the detail coefficients. Adam \cite{adam_optimizer} is employed to optimize the proposed three-staged loss function.

Finally, referring to Fig. \ref{proposed_network_architecture}, the SWANencoder's output gives the estimated abundance vector at each pixel location ($\boldsymbol{\alpha}_{i(e\times 1)}$). This output is directly used to construct the estimated abundance matrix $\mathbf{\widehat{A}}$ since the fractional contribution of each endmember remains consistent in the wavelet space with respect to the data acquisition domain. The connected weights of the SWANdecoder subnetwork represent $\{\mathbf{\widehat{M}_a}, \mathbf{\widehat{M}_d}\}$, i.e. the estimated set of approximation and detail coefficients of the endmember matrix. These are used to construct estimated endmember matrix $\mathbf{\widehat{M}}$ using the inverse DWT (IDWT) as $ \mathbf{\widehat{M}}_{L\times e} = IDWT(\{\mathbf{\widehat{M}_a}, \mathbf{\widehat{M}_d}\}_{K\times e})$.

\begin{figure}[t]
\includegraphics[width=8.5cm,height=6cm]{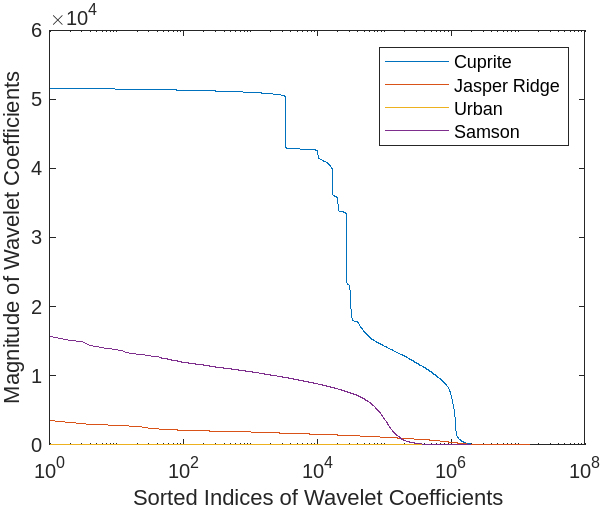}

\caption{Compressibility of hyperspectral imagery AVIRIS Cuprite \cite{datasets}, HYDICE Urban \cite{datasets}, Samson \cite{datasets}, and AVIRIS Jasper Ridge \cite{datasets}: The images are represented using \textit{bior 3.3} wavelet at optimum nodes of decomposition. It can be observed that \textit{bior 3.3} coefficients optimally decay as per the power law \cite{candes2005signal}. This helps represent hyperspectral images in a compact form.}
\label{bior_figure}
\end{figure}

\begin{table*}[t]
\caption{Data and Model Complexity Comparison for Different Hyperspectral Image Data Sets.}
\label{theoretical_analysis}
\centering
\begin{tabular}{ |M{1.7cm}|M{2.5cm}|M{2cm}|M{2cm}|M{2cm}|M{2cm}|M{2cm}|} \hline
\multicolumn{2}{|c|}{\textbf{Data}} & {\textbf{Synthetic HSI Data 1}} & {\textbf{Synthetic HSI Data 2}} & {\textbf{AVIRIS Jasper Ridge}} & {\textbf{Samson}} & {\textbf{HYDICE Urban}}\\ \hline
\multicolumn{2}{|c|}{\textbf{Image dimensions}} & 75$\times$75 & 128$\times$128 & 100$\times$100 & 95$\times$95 & 307$\times$307 \\ \hline
\multicolumn{2}{|c|}{\textbf{Given \# bands (L)}} & 224 & 431 & 224 & 156 & 210\\ \hline
\multicolumn{2}{|c|}{\textbf{\# bands with wavelet basis (K)}} & 115 & 219 & 102 & 81 & 84\\ \hline
\multicolumn{2}{|c|}{\textbf{\# endmembers (e)}} & 3 & 5 & 4 & 3 & 4\\ \hline
\textbf{\# trainable} & PA-HSU \cite{ssl_hsu} & 95,929 & 156,656 & 96,174 & 76,277 & 92,114 \\ \cline {2-7}
\textbf{parameters} & CNNAEU \cite{palsson2020convolutional} & 178,326 & 447,293 & 205,480 & 124,266 & 192,656 \\ \cline {2-7}
 & MiSiCNet \cite{rasti2022misicnet} & 1,715,765 & 2,199,618 & 1,718,296 & 1,558,617 & 1,685,928 \\ \cline {2-7}
 & proposed SWAN & \textbf{55,971} & \textbf{151,371} & \textbf{50,007} & \textbf{36,591} & \textbf{39,855} \\ \hline
\end{tabular}
\end{table*}

\section{Experimental Results}

We first conduct experiments on two benchmark synthetic hyperspectral image cubes constructed using real spectral signatures from the USGS spectral library \cite{usgs_dataset}. To assess the noise sensitivity, we have added different levels of white Gaussian noise to the image data sets and compared the results with state-of-the-art approaches. Next, we conduct experiments on three benchmark real hyperspectral image data sets, i.e., AVIRIS Jasper Ridge \cite{datasets}, HYDICE Urban \cite{datasets} and  Samson \cite{datasets}, to demonstrate the efficacy of the proposed method. We compare our results with relevant state-of-the-art neural network based unmixing approaches for which the codes are available in the public domain \cite{rasti2023image}. We have chosen the following approaches for comparison: EndNet \cite{endnet}, uDAS \cite{qu2018udas}, Wavelet AE \cite{wavelet_autoencoder}, PA-HSU \cite{ssl_hsu}, CNNAEU \cite{palsson2020convolutional}, MiSiCNet \cite{rasti2022misicnet}, and EDAA \cite{zouaoui2023edaa}. In all the experiments, the number of endmembers \textit{e} is considered using \cite{virtual_dim}.

\subsection{General Methodology}

Network training and inference are implemented in Python and Tensorflow 2.0 using Google Colaboratory. Before training, we split the \textit{P} input pixel vectors in the ratio of 80:20 between the train and the test sets, i.e. we randomly select 80\% of the pixels for training and the remaining 20\% for testing. To achieve stable training and consistent output, we normalize the DWT coefficients $\mathbf{x_w}$ pixel-wise at the input layer $l_0$ and use different batch sizes for each hyperspectral image data set. The optimum values of the regularisation parameters are empirically set to $\lambda_1=0.1$ and $\lambda_2=0.01$ using \cite{bhatt2016regularization}.  We train the complete network for 100 epochs using the three-staged loss function described in Eqs. (\ref{loss_function} - \ref{l9}) with Adam \cite{adam_optimizer} as the optimizer.

Once the training is complete, we freeze SWAN's weights during the inference to calculate $\mathbf{\widehat{M}}$ and $\mathbf{\widehat{A}}$. Referring to Fig. \ref{proposed_network_architecture}, we extract the SWANdecoder's weights as $\{\mathbf{\widehat{M}_a}, \mathbf{\widehat{M}_d}\}$ to construct $\mathbf{\widehat{M}}$ as $ \mathbf{\widehat{M}}_{L\times e} = IDWT(\{\mathbf{\widehat{M}_a}, \mathbf{\widehat{M}_d}\}_{K\times e})$. Finally, we use all the pixel vectors to extract $\boldsymbol{\alpha}_{i(e\times 1)}$ from SWANencoder's output to construct $\mathbf{\widehat{A}}$.

\subsection{Experiment on two synthetic hyperspectral image sets}

\begin{figure}[t]
\centering
\includegraphics[width=3.5in]{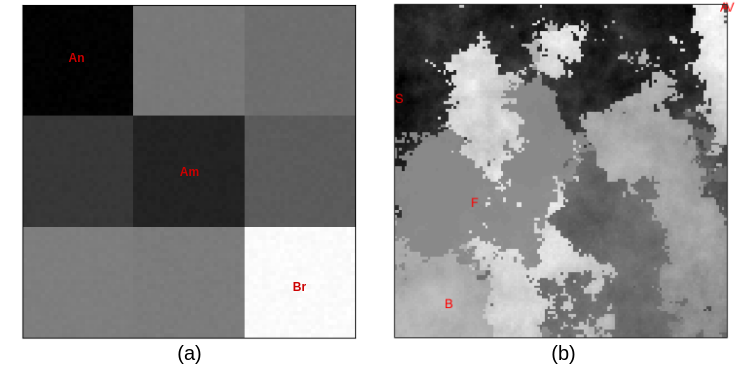}
\caption{Endmembers and their respective
locations are marked on the mean image: (a) Synthetic HSI data 1: Am-Ammonioalunite, An-Andradite, Br-Brucite, and (b) Synthetic HSI data 2: A-Asphalt, B-Brick, F-Fiberglass, S-Sheetmetal, V-Vinylplastic.}
\label{mean_data}
\end{figure}

\begin{table*}[t]
\caption{Average Error Scores by Different Algorithms for Synthetic Hyperspectral Data 1.}
\label{synthetic_endmembers}
\centering
\begin{tabular}{ |M{3cm}|M{0.85cm}|M{0.85cm}|M{0.85cm}|M{0.85cm}|M{0.85cm}|M{0.85cm}|M{0.85cm}|M{0.85cm}|M{0.85cm}|M{0.85cm}|M{0.85cm}|M{0.85cm}| } \hline
Data & \multicolumn{12}{c|}{\textbf{Synthetic Hyperspectral Data 1}}\\ \hline

Unmixed component & \multicolumn{12}{c|}{\textbf{Estimated endmembers}}\\ \hline
Performance metrics & \multicolumn{4}{c|}{\textbf{RMSE \cite{rmse_cite}}} & \multicolumn{4}{c|}{\textbf{SAD \cite{sad_cite}}} & \multicolumn{4}{c|}{\textbf{SID \cite{sid_cite}}}\\ \hline
SNR of data & 40 \textit{dB} & 20 \textit{dB} & 10 \textit{dB} & 5 \textit{dB} & 40 \textit{dB} & 20 \textit{dB} & 10 \textit{dB} & 5 \textit{dB} & 40 \textit{dB} & 20 \textit{dB} & 10 \textit{dB} & 5 \textit{dB}\\ \hline
EndNet \cite{endnet} & 0.1022 & 0.2344 & 1.3418 & 2.0031 & 1.3263 & 1.4265 & 2.6428 & 2.8922 & 0.0102 & 0.0112 & 0.0114 & 0.0172\\ \hline
uDAS\cite{qu2018udas} & 0.1135 & 0.3824 & 3.2918 & 3.8123 & 1.6207 & 1.6851 & 2.8443 & 2.9905 & 0.0116 & 0.0124 & 0.0128 & 0.0162\\ \hline
Wavelet AE \cite{wavelet_autoencoder} & 0.0862 & 0.1201 & 1.4056 & 2.1027 & 1.2008 & 1.3126 & 1.9912 & 2.7522 & 0.0121 & 0.0134 & 0.0142 & 0.0152 \\ \hline
PA-HSU \cite{ssl_hsu} & 0.0634 & 0.1124 & 1.1142 & 1.7591 & 1.1105 & 1.2802 & 1.5023 & 2.4927 & 0.0091 & 0.0104 & 0.0109 & 0.0118\\ \hline
CNNAEU \cite{palsson2020convolutional} & 0.1003 & 0.2379 & 1.8053 & 3.0067 & 1.3409 & 1.8723 & 2.0083 & 2.9031 & 0.0133 & 0.0149 & 0.0166 & 0.0191\\ \hline
MiSiCNet \cite{rasti2022misicnet} & 0.0912 & 0.1428 & 1.0623 & 1.4599 & 1.3286 & 1.9092 & 1.6620 & 2.7107 & 0.0319 & 0.0391 & 0.0444 & 0.0512\\ \hline
EDAA \cite{zouaoui2023edaa} & 0.0512 & 0.1528 & 1.0021 & 1.2967 & 1.1164 & 1.2528 & 1.3391 & 1.9522 & 0.0103 & 0.0219 & 0.0298 & 0.0311\\ \hline
proposed SWAN & \textbf{0.0242} & \textbf{0.0334} & \textbf{0.6210} & \textbf{0.9913} & \textbf{0.3224} & \textbf{0.5404} & \textbf{0.6173} & \textbf{1.1104} & \textbf{0.0032} & \textbf{0.0043} & \textbf{0.0068} & \textbf{0.0071}\\ \hline \hline

Unmixed component & \multicolumn{12}{c|}{\textbf{Estimated abundances}}\\ \hline
Performance metrics & \multicolumn{4}{c|}{\textbf{RMSE \cite{rmse_cite}}} & \multicolumn{4}{c|}{\textbf{SAD \cite{sad_cite}}} & \multicolumn{4}{c|}{\textbf{SID \cite{sid_cite}}}\\ \hline
SNR of data & 40 \textit{dB} & 20 \textit{dB} & 10 \textit{dB} & 5 \textit{dB} & 40 \textit{dB} & 20 \textit{dB} & 10 \textit{dB} & 5 \textit{dB} & 40 \textit{dB} & 20 \textit{dB} & 10 \textit{dB} & 5 \textit{dB}\\ \hline
EndNet \cite{endnet} & 0.1021 & 0.2021 & 0.2551 & 0.2941 & 0.3001 & 0.3108 & 0.3882 & 0.4226 & 1.0102 & 1.4028 & 1.5366 & 2.1771\\ \hline
uDAS\cite{qu2018udas} & 0.1124 & 0.2241 & 0.3007 & 0.4271 & 0.3234 & 0.3327 & 0.4523 & 0.5133 & 1.0841 & 1.4634 & 1.4988 & 2.3721\\ \hline
Wavelet AE \cite{wavelet_autoencoder} & 0.1013 & 0.1824 & 0.2285 & 0.2616 & 0.3128 & 0.3246 & 0.3393 & 0.3982 & 1.0162 & 1.2799 & 1.3959 & 2.0018\\ \hline
PA-HSU \cite{ssl_hsu} & 0.1006 & 0.1728 & 0.2198 & 0.2430 & 0.2742 & 0.2884 & 0.3087 & 0.3467 & 0.9240 & 1.0982 & 1.1027 & 1.64\\ \hline
CNNAEU \cite{palsson2020convolutional} & 0.1023 & 0.2023 & 0.2531 & 0.2833 & 0.3012 & 0.3123 & 0.3891 & 0.4231 & 1.2183 & 1.5031 & 1.8371 & 2.1781\\ \hline
MiSiCNet \cite{rasti2022misicnet} & 0.0891 & 0.1522 & 0.2012 & 0.2615 & 0.2591 & 0.3007 & 0.3164 & 0.3492 & 1.0298 & 1.1145 & 1.1982 & 1.2172\\ \hline
EDAA \cite{zouaoui2023edaa} & 0.0922 & 0.1462 & 0.2139 & 0.2911 & 0.2663 & 0.3068 & 0.3225 & 0.3312 & 0.9138 & 1.1074 & 1.1135 & 1.4957\\ \hline
proposed SWAN & \textbf{0.0426} & \textbf{0.1029} & \textbf{0.1723} & \textbf{0.2002} & \textbf{0.1274} & \textbf{0.1771} & \textbf{0.2009} & \textbf{0.2491} & \textbf{0.6150} & \textbf{1.0061} & \textbf{1.0128} & \textbf{1.1137}\\ \hline
\end{tabular}
\end{table*}

\begin{figure*}[t]
\centering
\includegraphics[width=\textwidth]{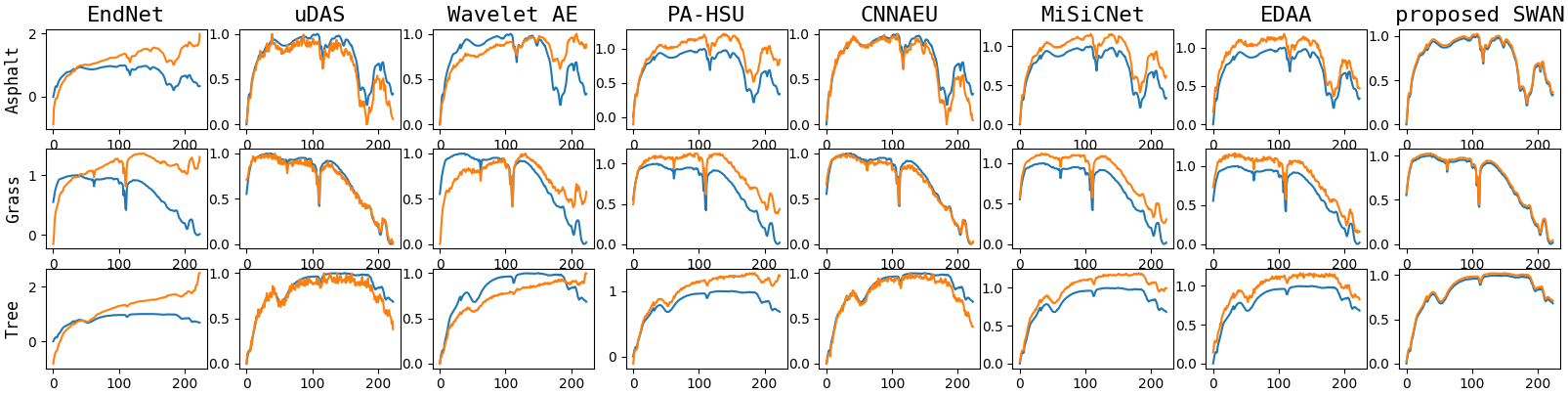}
\caption{Extracted endmembers for synthetic data 1 at 10 \textit{dB} SNR. Ground truth endmembers are displayed in blue while their estimates are in orange.}
\label{checkerboard_endmembers}
\end{figure*}

\subsubsection{Synthetic Hyperspectral Image Data 1}

Three real endmember signatures are selected from the USGS digital spectral library, viz. Ammonioalunite, Brucite, and Andradite spread in the 224 bands ranging from 400 \textit{nm} - 2500 \textit{nm}. Arranging each endmember in a column vector forms the endmember matrix $\mathbf{M}$ of size $224 \times 3$. Now, considering three different spatial patterns, three abundance maps each of size $75 \times 75$ pixels are constructed. These endmembers and abundances synthesize a 224-band ground truth hyperspectral image data cube, i.e. $75 \times 75 \times 224$. For noise sensitivity analysis, independent and identically distributed Gaussian noises are added at different proportions. Referring to the discussion in Section \rom{4}, we use \textit{bior3.3} family of wavelet bases for DWT, since it best characterizes the hyperspectral imagery by capturing the variations compactly and completely. It yields a sparse representation of the image cube with size $75 \times 75 \times 115$. The mean image of the synthetic data set along with marked endmembers is shown in Fig. \ref{mean_data}(a).

Next, we split the 5,625 pixel vectors in the ratio of 80:20 as the train-test split. We then train the network with a batch size of 50. We train the complete network using the three-staged loss function described in Eqs. (\ref{loss_function} - \ref{l9}). Finally, we use all the vectors during inference to calculate $\mathbf{\widehat{A}}$ from SWANencoder's output and $\mathbf{\widehat{M}}$ from SWANdecoder's weights.

\begin{figure}[t]
\centering
\includegraphics[width=3.5in]{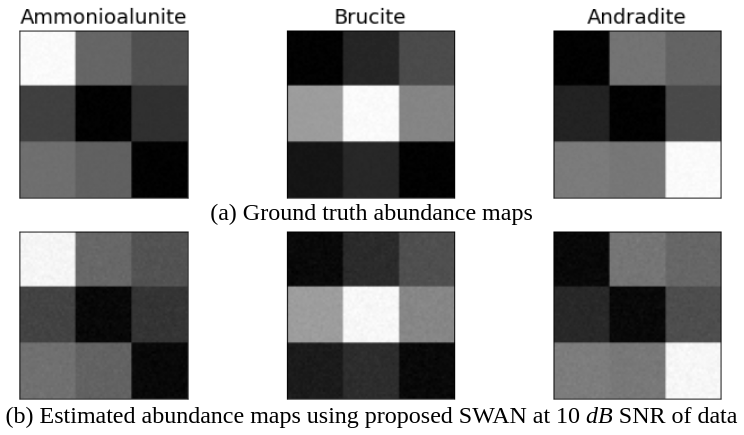}
\caption{Estimated abundance maps for synthetic data 1 at 10 \textit{dB} SNR.}
\label{checkerboard_abundances}
\end{figure}

The signatures of three endmembers are estimated using SWAN and compared with state-of-the-art. A result at 10 \textit{dB} SNR is shown in Fig. \ref{checkerboard_endmembers}. It can be seen that the estimated endmembers are almost overlapped with the ground truth endmembers. The estimated abundance maps displayed in Fig. \ref{checkerboard_abundances} reveal visually consistent spatial patterns of estimated abundances using proposed SWAN even at 10 \textit{dB} SNR in data. Table \ref{synthetic_endmembers} provides comparative average error scores in terms of root-mean-squared error (RMSE) \cite{rmse_cite}, spectral angle
distance (SAD) \cite{sad_cite}, and spectral information divergence (SID) \cite{sid_cite}. It confirms the robustness of SWAN for HSIU even at high noise levels in the image.

\subsubsection{Synthetic Hyperspectral Image Data 2}

\begin{table*}[t]
\caption{Average Error Scores by Different Algorithms for Synthetic Hyperspectral Data 2.}
\label{spherical_results}
\centering
\begin{tabular}{ |M{3cm}|M{0.85cm}|M{0.85cm}|M{0.85cm}|M{0.85cm}|M{0.85cm}|M{0.85cm}|M{0.85cm}|M{0.85cm}|M{0.85cm}|M{0.85cm}|M{0.85cm}|M{0.85cm}| } \hline
Data & \multicolumn{12}{c|}{\textbf{Synthetic Hyperspectral Data 2}}\\ \hline

Unmixed component & \multicolumn{12}{c|}{\textbf{Estimated endmembers}}\\ \hline
Performance metrics & \multicolumn{4}{c|}{\textbf{RMSE \cite{rmse_cite}}} & \multicolumn{4}{c|}{\textbf{SAD \cite{sad_cite}}} & \multicolumn{4}{c|}{\textbf{SID \cite{sid_cite}}}\\ \hline
SNR of data & 60 \textit{dB} & 40 \textit{dB} & 20 \textit{dB} & 10 \textit{dB} & 60 \textit{dB} & 40 \textit{dB} & 20 \textit{dB} & 10 \textit{dB} & 60 \textit{dB} & 40 \textit{dB} & 20 \textit{dB} & 10 \textit{dB}\\ \hline
EndNet \cite{endnet} & 0.1243 & 0.2974 & 1.2235 & 1.9822 & 1.2413 & 1.5672 & 2.3426 & 2.7940 & 0.0237 & 0.0322 & 0.0396 & 0.0402\\ \hline
uDAS\cite{qu2018udas} & 0.1248 & 0.5918 & 2.3905 & 3.0327 & 1.5106 & 1.8843 & 2.4903 & 2.8012 & 0.0286 & 0.0299 & 0.0352 & 0.0388\\ \hline
Wavelet AE \cite{wavelet_autoencoder} & 0.1068 & 0.3168 & 1.7644 & 2.2974 & 1.1013 & 1.4677 & 2.0344 & 2.9742 & 0.0237 & 0.0298 & 0.0311 & 0.0372 \\ \hline
PA-HSU \cite{ssl_hsu} & 0.0718 & 0.1294 & 1.2374 & 1.9108 & 1.1674 & 1.2317 & 1.4977 & 2.0374 & 0.0103 & 0.0167 & 0.0241 & 0.0289\\ \hline
CNNAEU \cite{palsson2020convolutional} & 0.1372 & 0.4316 & 1.6853 & 2.3973 & 1.3982 & 1.9653 & 2.6171 & 2.8876 & 0.0313 & 0.0472 & 0.0570 & 0.0718\\ \hline
MiSiCNet \cite{rasti2022misicnet} & 0.1106 & 0.2184 & 1.3727 & 1.8622 & 1.0121 & 1.3372 & 1.6022 & 1.9913 & 0.0255 & 0.0313 & 0.0479 & 0.0512\\ \hline
EDAA \cite{zouaoui2023edaa} & 0.0822 & 0.1344 & 1.1928 & 1.7352 & 0.9874 & 1.1829 & 1.5899 & 1.8722 & 0.0087 & 0.0123 & 0.0189 & 0.0212\\ \hline
proposed SWAN & \textbf{0.0438} & \textbf{0.0743} & \textbf{0.5312} & \textbf{0.9133} & \textbf{0.4012} & \textbf{0.5777} & \textbf{0.7644} & \textbf{1.0388} & \textbf{0.0013} & \textbf{0.0037} & \textbf{0.0066} & \textbf{0.0084}\\ \hline \hline

Unmixed component & \multicolumn{12}{c|}{\textbf{Estimated abundances}}\\ \hline
Performance metrics & \multicolumn{4}{c|}{\textbf{RMSE \cite{rmse_cite}}} & \multicolumn{4}{c|}{\textbf{SAD \cite{sad_cite}}} & \multicolumn{4}{c|}{\textbf{SID \cite{sid_cite}}}\\ \hline
SNR of data & 60 \textit{dB} & 40 \textit{dB} & 20 \textit{dB} & 10 \textit{dB} & 60 \textit{dB} & 40 \textit{dB} & 20 \textit{dB} & 10 \textit{dB} & 60 \textit{dB} & 40 \textit{dB} & 20 \textit{dB} & 10 \textit{dB}\\ \hline
EndNet \cite{endnet} & 0.1137 & 0.1978 & 0.2833 & 0.3367 & 0.2374 & 0.2966 & 0.3020 & 0.3111 & 1.0034 & 1.5794 & 1.5564 & 2.3451\\ \hline
uDAS\cite{qu2018udas} & 0.1074 & 0.2137 & 0.4678 & 0.5041 & 0.2035 & 0.2674 & 0.3341 & 0.3977 & 1.0034 & 1.5670 & 1.9944 & 2.2301\\ \hline
Wavelet AE \cite{wavelet_autoencoder} & 0.1304 & 0.1955 & 0.2341 & 0.2664 & 0.1642 & 0.1734 & 0.2994 & 0.3126 & 1.0244 & 1.3497 & 1.6477 & 2.1368\\ \hline
PA-HSU \cite{ssl_hsu} & 0.1047 & 0.1974 & 0.2034 & 0.2134 & 0.2674 & 0.2997 & 0.3105 & 0.3674 & 0.9944 & 1.0843 & 1.1677 & 1.8662\\ \hline
CNNAEU \cite{palsson2020convolutional} & 0.1274 & 0.2690 & 0.3016 & 0.4678 & 0.2911 & 0.3663 & 0.4610 & 0.5032 & 1.1023 & 1.4928 & 1.8017 & 2.4520\\ \hline
MiSiCNet \cite{rasti2022misicnet} & 0.1012 & 0.1800 & 0.2134 & 0.2691 & 0.1812 & 0.2064 & 0.3019 & 0.3266 & 1.0290 & 1.2248 & 1.4461 & 2.1059\\ \hline
EDAA \cite{zouaoui2023edaa} & 0.0994 & 0.1702 & 0.2011 & 0.2105 & 0.1544 & 0.1833 & 0.2902 & 0.3001 & 0.9877 & 1.1677 & 1.3017 & 1.7542\\ \hline
proposed SWAN & \textbf{0.0755} & \textbf{0.1167} & \textbf{0.1849} & \textbf{0.2034} & \textbf{0.1334} & \textbf{0.1849} & \textbf{0.2016} & \textbf{0.2511} & \textbf{0.5377} & \textbf{1.0674} & \textbf{1.0977} & \textbf{1.1084}\\ \hline
\end{tabular}
\end{table*}

\begin{figure*}[t]
\centering
\includegraphics[width=\textwidth, height=6cm]{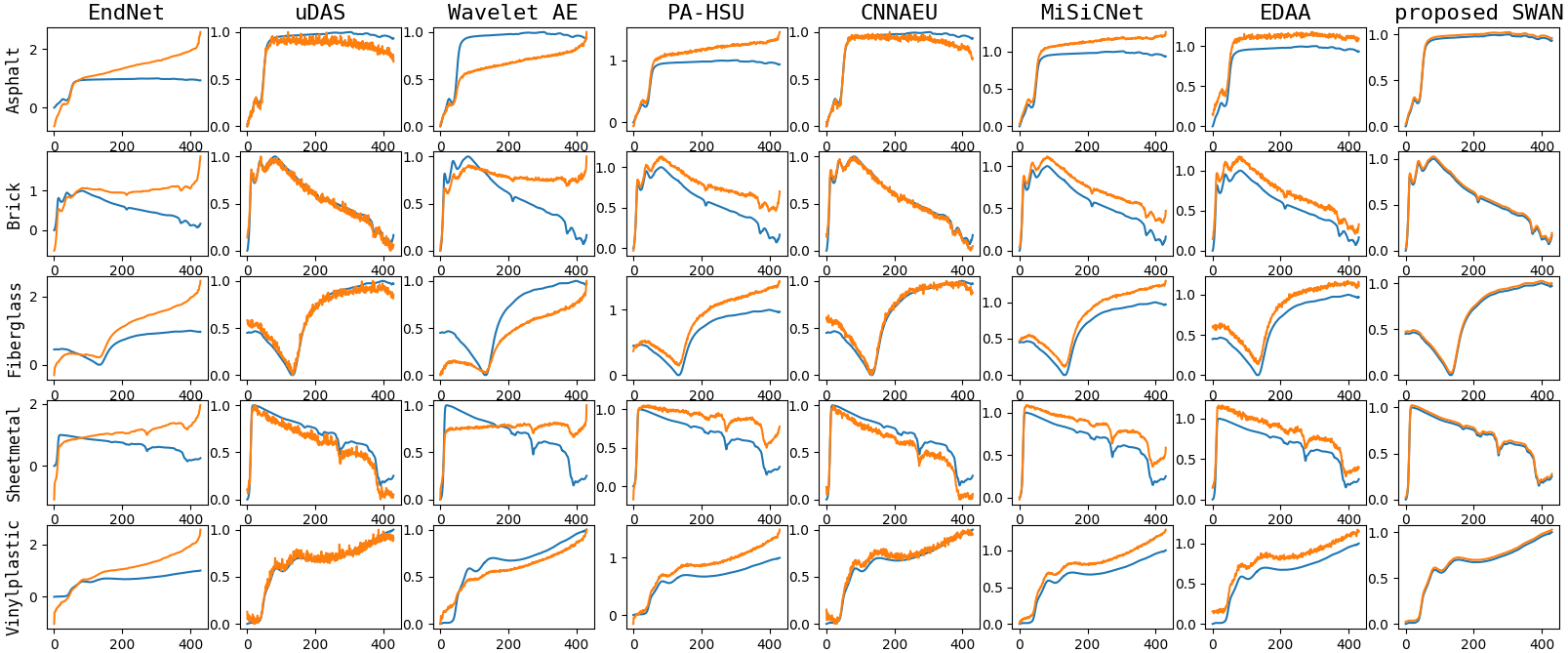}
\caption{Extracted endmembers for synthetic data 2 at 10 \textit{dB} SNR. Ground truth endmembers are displayed in blue while their estimates are in orange.}
\label{spherical_endmembers}
\end{figure*}

\begin{figure}[t]
\centering
\includegraphics[width=3.5in]{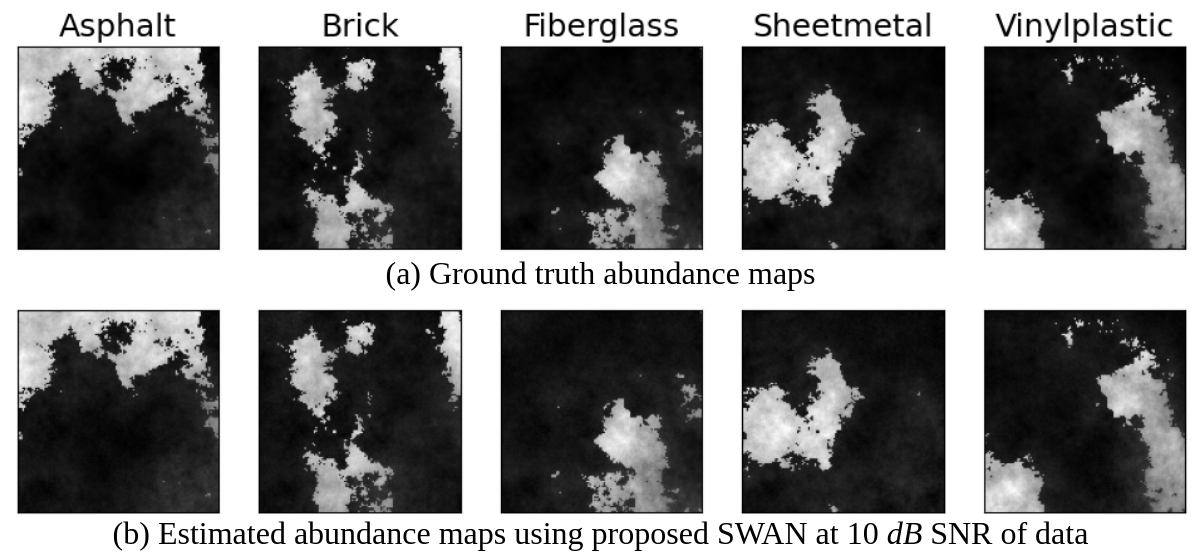}
\caption{Estimated abundance maps for synthetic data 2 at 10 \textit{dB} SNR.}
\label{spherical_abundances}
\end{figure}

Here, another five endmembers are selected from the USGS spectral library, viz. Asphalt, Brick, Fiberglass, Sheetmetal, and Vinylplastic spread in 431 bands ranging from 400 \textit{nm} - 2500 \textit{nm}. They form the endmember matrix $\mathbf{M}$ of size $431 \times 5$.  The spheric Gaussian field with different parameters is considered for generating abundance maps in the data set, as described in \cite{blind_hsu}. Considering three different abundance maps each of size $128 \times 128$, a 431-band ground truth hyperspectral image cube is synthesized, i.e. $128 \times 128 \times 431$.  The mean image of the synthetic data set along with marked endmembers is shown in Fig. \ref{mean_data}(b). For noise sensitivity analysis, independent and identically distributed Gaussian noises are added at different proportions. The use of \textit{bior3.3} family of wavelet bases for DWT \cite{blind_hsu} yields a sparse representation of the image cube with size $128 \times 128 \times 219$.

For training and testing SWAN, we split the 16,384 pixel vectors in the ratio of 80:20 as the train-test split and train the network with a batch size of 50.  The signatures of five endmembers are estimated using SWAN and compared with state-of-the-art. A result of extracted endmembers at 10 \textit{dB} SNR in the data is shown in Fig. \ref{spherical_endmembers}. It can be seen that the estimated endmembers using proposed SWAN have a close match with the ground truth signatures. The corresponding estimated abundance maps shown in Fig. \ref{spherical_abundances} show visually consistent spatial patterns of abundances using proposed SWAN even at 10 \textit{dB} SNR in data. Table \ref{spherical_results} lists comparative average error scores for endmembers and abundances in terms of RMSE \cite{rmse_cite}, SAD \cite{sad_cite}, and spectral information divergence (SID) \cite{sid_cite}. It further quantifies the superiority of the proposed SWAN when compared to state-of-the-art methods.

Both qualitative and quantitative results in experiments with synthetic images, combined with the model complexity shown in Table \ref{theoretical_analysis}, indicate the capabilities of proposed SWAN as a compact, efficient, and robust method for HSIU.

\subsection{Experiment on three real hyperspectral image data sets}
\begin{table*}[h!]
\caption{Average Error Scores by Different Algorithms for AVIRIS Jasper Ridge Hyperspectral Image Cube.}
\label{jasper_results}
\centering
\begin{tabular}{ |M{3.25cm}|M{1.75cm}|M{1.75cm}|M{1.75cm}|M{1.75cm}|M{1.75cm}|M{1.75cm}| }
\hline
Data & \multicolumn{6}{c|}{\textbf{AVIRIS Jasper Ridge}}\\ \hline
Unmixing component & \multicolumn{3}{c|}{\textbf{Estimated endmembers}} & \multicolumn{3}{c|}{\textbf{Estimated abundances}}\\ \hline
Performance metrics & \textbf{RMSE \cite{rmse_cite}} & \textbf{SAD \cite{sad_cite}} & \textbf{SID \cite{sid_cite}} & \textbf{RMSE \cite{rmse_cite}} & \textbf{SAD \cite{sad_cite}} & \textbf{SID \cite{sid_cite}}\\ \hline
EndNet \cite{endnet} & 8.6352 & 1.0121 & 0.0396 & 0.0041 & 2.3479 & 2.8921\\ \hline
uDAS \cite{qu2018udas} & 7.0542 & 1.7744 & 0.0315 & 0.0028 & 2.0110 & 1.7423\\ \hline
Wavelet AE \cite{wavelet_autoencoder} & 2.8120 & 2.7192 & 0.0305 & 0.0025 & 2.1173 & 1.6702\\ \hline
PA-HSU \cite{ssl_hsu} & 2.4067 & 2.4056 & 0.0235 & 0.0017 & 2.4598 & 1.6713\\ \hline
CNNAEU \cite{palsson2020convolutional} & 5.1982 & 2.9201 & 0.0428 & 0.0128 & 2.7102 & 2.0012\\ \hline
MiSiCNet \cite{rasti2022misicnet} & 3.7019 & 2.2123 & 0.0329 & 0.0032 & 2.1098 & 1.9813\\ \hline
EDAA \cite{zouaoui2023edaa} & 2.2116 & 1.1194 & 0.0206 & 0.0021 & 1.8012 & 1.7723\\ \hline
proposed SWAN & \textbf{1.9176} & \textbf{0.6228} & \textbf{0.0218} & \textbf{0.0011} & \textbf{1.2013} & \textbf{1.6629}\\ \hline
\end{tabular}
\end{table*}

\begin{table*}[h!]
\caption{Average Error Scores by Different Algorithms for HYDICE Urban Hyperspectral Image Cube.}
\label{urban_results}
\centering
\begin{tabular}{ |M{3.25cm}|M{1.75cm}|M{1.75cm}|M{1.75cm}|M{1.75cm}|M{1.75cm}|M{1.75cm}| }
\hline
Data & \multicolumn{6}{c|}{\textbf{HYDICE Urban}}\\ \hline
Unmixing component & \multicolumn{3}{c|}{\textbf{Estimated endmembers}} & \multicolumn{3}{c|}{\textbf{Estimated abundances}}\\ \hline
Performance metrics & \textbf{RMSE \cite{rmse_cite}} & \textbf{SAD \cite{sad_cite}} & \textbf{SID \cite{sid_cite}} & \textbf{RMSE \cite{rmse_cite}} & \textbf{SAD \cite{sad_cite}} & \textbf{SID \cite{sid_cite}}\\ \hline
EndNet \cite{endnet} & 9.6301 & 1.8436 & 0.0721 & 0.0972 & 0.9915 & 2.7109\\ \hline
uDAS \cite{qu2018udas} & 7.4420 & 1.7749 & 0.0492 & 0.0920 & 0.9316 & 2.3024\\ \hline
Wavelet AE \cite{wavelet_autoencoder} & 3.2366 & 1.5581 & 0.0329 & 0.0891 & 0.9126 & 2.9082\\ \hline
PA-HSU \cite{ssl_hsu} & 2.9812 & 1.3201 & 0.0310 & 0.0794 & 0.8018 & 1.9921\\ \hline
CNNAEU \cite{palsson2020convolutional} & 8.7201 & 2.0913 & 0.0673 & 0.1026 & 1.0347 & 2.8102\\ \hline
MiSiCNet \cite{rasti2022misicnet} & 2.1022 & 1.2726 & 0.0418 & 0.0813 & 0.7217 & 1.9135\\ \hline
EDAA \cite{zouaoui2023edaa} & 1.9018 & 0.8170 & 0.0521 & 0.0752 & 0.6086 & 1.8177\\ \hline
proposed SWAN & \textbf{1.1692} & \textbf{0.6017} & \textbf{0.0303} & \textbf{0.0604} & \textbf{0.4960} & \textbf{1.6033}\\ \hline
\end{tabular}
\end{table*}

\begin{table*}[h!]
\caption{Average Error Scores by Different Algorithms for Samson Hyperspectral Image Cube.}
\label{samson_results}
\centering
\begin{tabular}{ |M{3.25cm}|M{1.75cm}|M{1.75cm}|M{1.75cm}|M{1.75cm}|M{1.75cm}|M{1.75cm}| }
\hline
Data & \multicolumn{6}{c|}{\textbf{Samson}}\\ \hline
Unmixing component & \multicolumn{3}{c|}{\textbf{Estimated endmembers}} & \multicolumn{3}{c|}{\textbf{Estimated abundances}}\\ \hline
Performance metrics & \textbf{RMSE \cite{rmse_cite}} & \textbf{SAD \cite{sad_cite}} & \textbf{SID \cite{sid_cite}} & \textbf{RMSE \cite{rmse_cite}} & \textbf{SAD \cite{sad_cite}} & \textbf{SID \cite{sid_cite}}\\ \hline
EndNet \cite{endnet} & 1.0213 & 2.1658 & 0.0035 & 0.0189 & 1.8371 & 0.9123\\ \hline
uDAS \cite{qu2018udas} & 0.7270 & 1.9342 & 0.0030 & 0.0145 & 2.2560 & 0.8012\\ \hline
Wavelet AE \cite{wavelet_autoencoder} & 0.0212 & 1.5017 & 0.0029 & 0.0130 & 1.7124 & 0.7129\\ \hline
PA-HSU \cite{ssl_hsu} & 0.0247 & 1.6920 & 0.0026 & 0.0126 & 1.9492 & 0.5858\\ \hline
CNNAEU \cite{palsson2020convolutional} & 2.1038 & 2.2288 & 0.0125 & 0.0472 & 1.6679 & 0.8238\\ \hline
MiSiCNet \cite{rasti2022misicnet} & 1.7652 & 1.8304 & 0.0098 & 0.0258 & 1.5612 & 0.7321\\ \hline
EDAA \cite{zouaoui2023edaa} & 0.7024 & 0.9821 & 0.0043 & 0.0141 & 1.4033 & 0.6982\\ \hline
proposed SWAN &  \textbf{0.0160} & \textbf{0.2482} & \textbf{0.0019} & \textbf{0.0118} & \textbf{1.1857} & \textbf{0.5620}\\ \hline
\end{tabular}
\end{table*}

\subsubsection{AVIRIS Jasper Ridge}

\begin{figure*}[t]
\centering
\includegraphics[width=\textwidth, height=5cm]{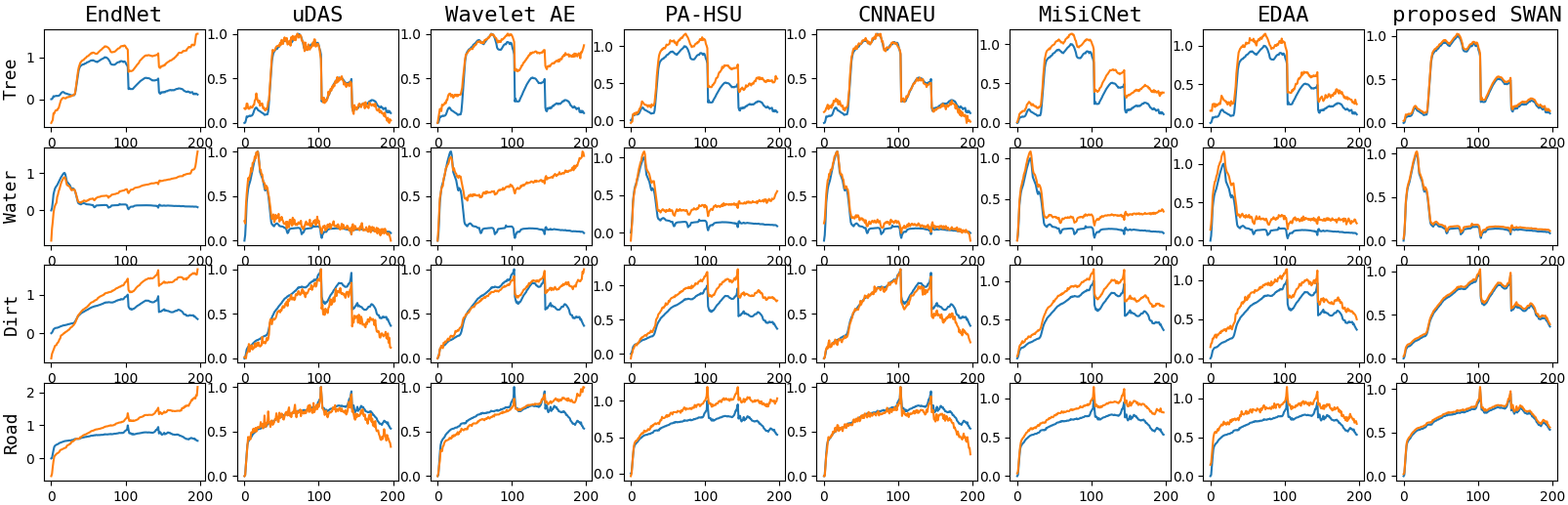}
\caption{Extracted endmembers for AVIRIS Jasper Ridge. Ground truth endmembers are displayed in blue while their estimates are in orange.}
\label{jasper_endmembers}
\end{figure*}

\begin{figure*}[t]
\centering
\includegraphics[width=\textwidth, height=5cm]{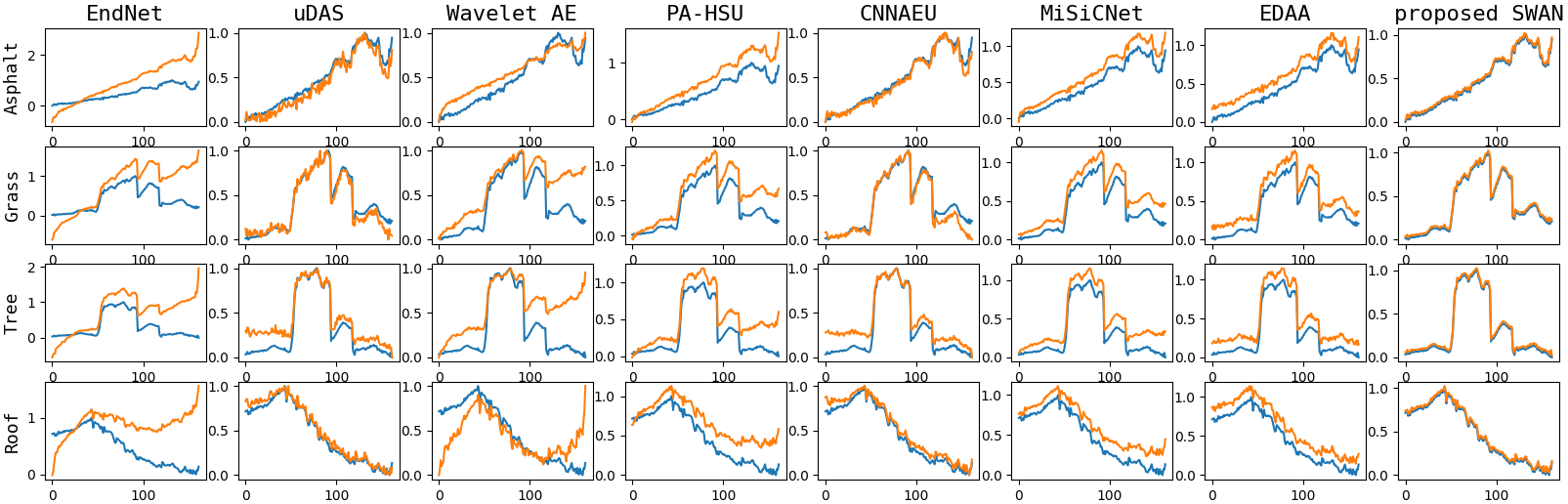}
\caption{Extracted endmembers for HYDICE Urban. Ground truth endmembers are displayed in blue while their estimates are in orange.}
\label{urban_endmembers}
\end{figure*}

\begin{figure*}[t]
\centering
\includegraphics[width=\textwidth]{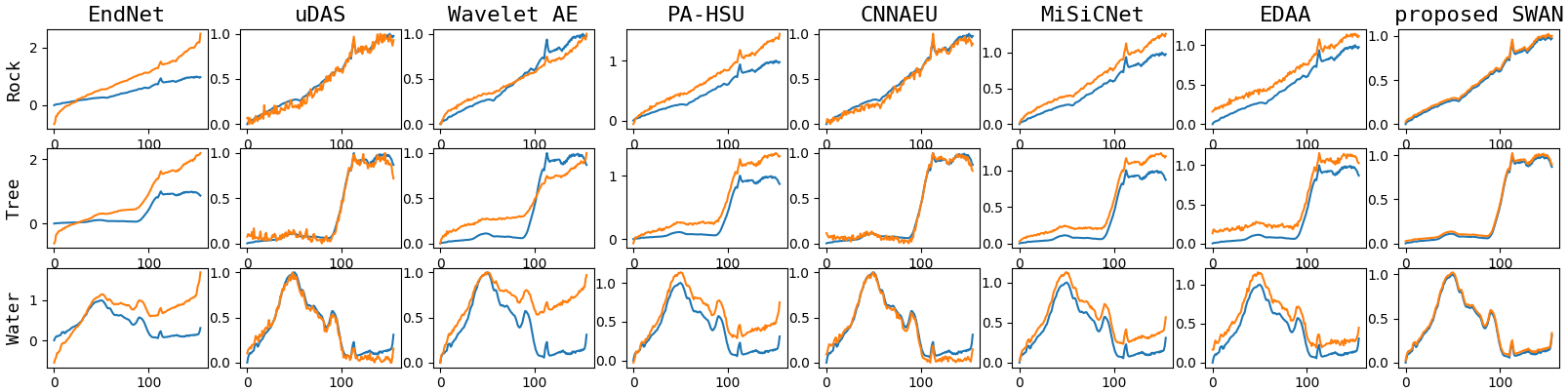}
\caption{Extracted endmembers by different approaches for Samson. Ground truth endmembers are displayed in blue while their estimates are in orange.}
\label{samson_endmembers}
\end{figure*}

\begin{figure}[t]
\centering
\includegraphics[width=3.5in]{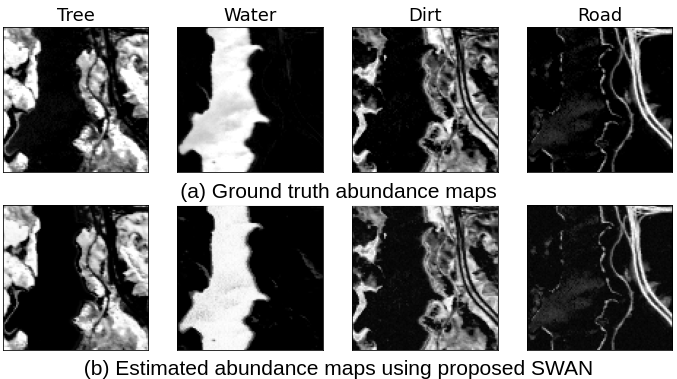}
\caption{Estimated abundance maps for AVIRIS Jasper Ridge.}
\label{jasper_abundances}
\end{figure}

\begin{figure}[t]
\centering
\includegraphics[width=3.5in]{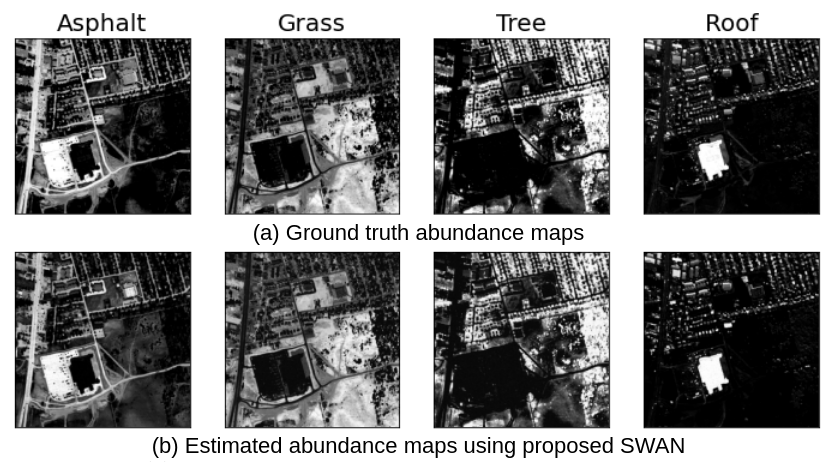}
\caption{Estimated abundance maps for HYDICE Urban.}
\label{urban_abundances}
\end{figure}

\begin{figure}[t]
\centering
\includegraphics[width=8cm]{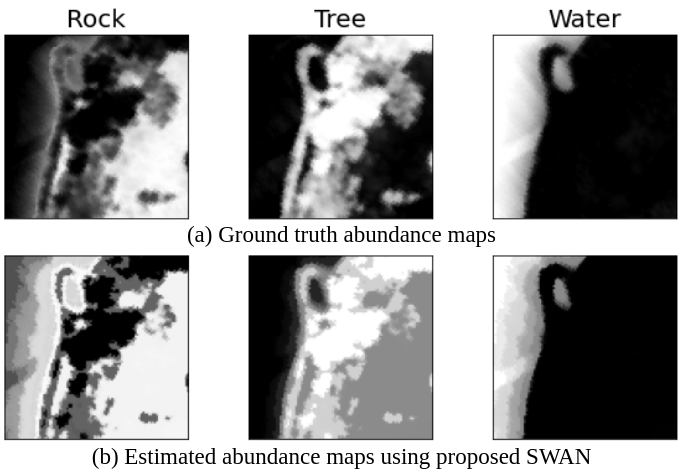}
\caption{Estimated abundance maps for Samson.}
\label{samson_abundances}
\end{figure}

We first use the real hyperspectral data collected by the AVIRIS sensor at Jasper Ridge \cite{datasets}. This is acquired in 224 contiguous wavelength channels starting from 380 \textit{nm} to 2500 \textit{nm} with up to 9.46 \textit{nm} spectral resolution. In this work, we consider a part image of $100 \times 100$ pixels where four land cover types (number of endmembers) are observed: Tree, Water, Dirt, and Road. For our experiment, the images are first represented in the wavelet domain as a cube of size $100 \times 100 \times 102$. Endmembers estimated using the proposed SWAN are shown and compared with different methods in Fig. \ref{jasper_endmembers}, while Fig. \ref{jasper_abundances} shows the estimated abundance maps corresponding to each extracted endmember. Table \ref{jasper_results} lists average error scores of estimated endmembers and abundance values obtained by the proposed SWAN concerning the ground truth as well as when compared with state-of-the-art. It quantitatively confirms the superiority of the proposed network.

\subsubsection{HYDICE Urban}

Next, the proposed SWAN is evaluated on a second real HYDICE Urban hyperspectral image \cite{datasets}. It includes 210 bands with a spectral resolution of 10\textit{nm}, out of which 162 bands are retained by removing some bands of dense water vapour and atmospheric effects. Each image size then results in $307 \times 307$ pixels, having four endmmebers (land cover types): viz, Asphalt, Grass, Tree, and Roof. The data cube is represented in the wavelet domain transforming it  as a cube of size $307 \times 307 \times 84$ in lieu of $307 \times 307 \times 162$. Corresponding results on endmembers are shown in Fig. \ref{urban_endmembers}, while Fig. \ref{urban_abundances} shows respective abundance maps for each land cover type against the ground truth. The quantitative analysis and comparison with state-of-the-art methods are listed in Table \ref{urban_results}.

\subsubsection{Samson}

Finally, the proposed SWAN is tested on another real Samson hyperspectral imagery \cite{datasets}. It includes 156 band images covering the wavelengths from 0.401 $\mu m$ to 0.889 $\mu m$, with each image size as $95 \times 95$ pixels. The site has found three major endmembers: Soil, Tree, and Water. We first represent the Samson image dataset of size $95 \times 95 \times 156$ into corresponding \textit{bior3.3} wavelet bases that gives a compact representation of size $95 \times 95 \times 81$. Corresponding qualitative and quantitative results are shown in Fig. \ref{samson_endmembers} (endmembers), Fig. \ref{samson_abundances} (abundance maps), and Table \ref{samson_results}.

The qualitative and quantitative results on both synthetic as well as the real data sets demonstrate the efficacy of the proposed SWAN. We learn the unmixing function from the compact, multi-scale wavelet representation to better exploit symmetries in terms of structures and invariances from wavelet coefficients. It further leverages incorporated kernel regularizers to better bound the magnitudes and helps preserve the variation in the estimated unmixed components.

\section{Conclusion}
This paper demonstrates a novel wavelet-based, self-supervised neural network for learning a blind unmixing function for hyperspectral images. The hierarchical non-linearity captures the latent symmetries as the compact wavelet representation. It enables the proposed method to exploit underlying covariant symmetries of wavelet coefficients in order to learn a better unmixing function. The proposed loss function further facilitates self-supervised learning and helps avoid the need for ground truth during model training. It incorporates kernel regularizers to bound the magnitudes and preserve spatial variations in the estimated endmember coefficients. Qualitative and quantitative result analysis and comparison with state-of-the-art neural network-based methods validate the efficacy of the proposed SWAN. One may design and deploy such compact neural models for resource-constrained unmixing.

\bibliographystyle{IEEEtran}
\bibliography{main}

\end{document}